%% file: main.tex
\documentclass{svproc}
\usepackage{url}
\usepackage{xcolor}
\usepackage{amsmath}
\usepackage{amsfonts}
\usepackage{amssymb}

\usepackage{graphicx}
\usepackage{subcaption}
\usepackage{multirow, multicol}
\usepackage{hyperref}
\usepackage{epstopdf}

\graphicspath{{images/}}

\newcommand{\e}[1]{\ensuremath{\boldsymbol{e}_{#1}}} 
\newcommand{\mv}[1]{\ensuremath{\boldsymbol{#1}}}

\begin{document}
\mainmatter   

\title{Inverse kinematic solution for generic 3R positional robots using Conformal Geometric Algebra}
\titlerunning{IKM for 3R positional robots with CGA} 
\author{Abhilash Nayak\inst{1} \and Durgesh Haribhau Salunkhe\inst{2}}
\authorrunning{Nayak A. and Salunkhe D. H.} 

\institute{
    \email{abhilash.un@outlook.com}
\and
    \email{salunkhedurgesh@outlook.com}
}

\maketitle           

\input{sections/0_abstract}
\input{sections/1_introduction}

\input{sections/2_CGA_basics}

\input{sections/3_kinematic_model}

\input{sections/4_conclusions}

%
\bibliographystyle{unsrt}
\bibliography{ref.bib}
\end{document}

%% file: sections/0_abstract.tex
\begin{abstract}
The inverse kinematics of generic 3R robots has been investigated through multiple approaches, mainly algebraic methods involving the solution of certain equation sets. Previous geometric interpretations of the solution, characterized as the intersection of a pair of conics have been confined to the joint-space domain. In this article, we study the Inverse Kinematic Model (IKM) of 3R robots, using the advantages of Conformal Geometric Algebra (CGA) to provide further insights on its kinematic properties. Our approach directly yields a univariate polynomial in terms of $\theta_2$ without the need to eliminate $\theta_1$ and $\theta_3$ by reframing the problem as the intersection of two circles, which are fundamental elements within this algebraic framework.

\keywords{kinematics, conformal geometric algebra, geometry}
\end{abstract}

%% file: sections/1_introduction.tex
\section{Introduction}
The inverse kinematics problem~(IKP) for a positional 3R robot was first addressed by Pieper~\cite{pieper_kinematics_1968} in 1968, who derived a univariate polynomial in $t_3 = \tan\frac{\theta_3}{2}$ and obtained $\theta_1$ and $\theta_2$ through backpropagation. However this method does not work for $a_1=0$ and $\alpha_1=0$ (refer Fig.~\ref{fig:ex_3R}). Selig later presented a compact solution using Lie algebra without such limitations, mentioning that inverse kinematics can be reduced to finding the intersection of a conic and a circle in the $\cos(\theta_1)\mbox{-}\sin(\theta_1)$ plane~\cite[Section 5.2]{selig}. This same geometric interpretation, applied to Pieper's approach but in the $\cos(\theta_3)\mbox{-}\sin(\theta_3)$ plane, was used to investigate the number of inverse kinematic solutions and cuspidal properties of the robot~\cite{wenger_cuspidal_2022}. These approaches discuss geometric interpretation of the obtained univariate polynomial after elimination. However, they fail to provide intuition about the kinematic model of the robot itself. In this article, we demonstrate how solving the problem using Conformal Geometric Algebra~(CGA) leads to an elimination free approach allowing a deeper understanding of inverse kinematic solutions in workspace and joint-space representations.\\
CGA provides a 5-dimensional representation of 3-dimensional Euclidean space \cite{Dorst2009,Lasenby2004}. This embedding offers two significant advantages for kinematic analysis: (i) rotations and translations can be unified as orthogonal transformations, and (ii) circles and spheres become fundamental elements of the algebra that can be manipulated and transformed in the same way as points, planes, and lines. To date, publications addressing inverse kinematics problems using CGA have primarily focused on non-generic serial robots (as defined in \cite{pai_genericity_1992}). A notable example is the inverse kinematics solution for anthropomorphic structures~\cite{Velasco2024,Lavor2018}. In the specific case of 3R serial chains, we note that the claims made in~\cite{Zaplana2022} cannot be generalized to generic 3R robots (specifically, in Equation~(70), $P_2$ does not lie on $\Pi_1$, $s_1$ is ill defined and $s_2$ is wrong).\\
In this article, we present a novel approach to solve IKM of a generic 3R robot using CGA. The method provides a geometric understanding of the IKM, allowing insight into the number of inverse kinematic solutions (IKS) and their distribution. The approach directly yields a univariate polynomial in $\theta_2$, thereby eliminating the need for algebraic manipulations to eliminate two of the three variables. Additionally, the method does not have degeneracy conditions, thus unifying the inverse kinematic model of 3R robots.

%% file: sections/2_CGA_basics.tex
\section{CGA: Notations and basic operations}
\label{sec:cga_basics}
The 5-dimensional conformal geometric algebra $\mathbb{G}_{4,1}$ is described with the orthogonal unit vectors $\e{i}^2 = +1$ for $i=1,...,4$ and $\e{5}^2 = -1$. 
A fundamental operation in geometric algebra is the geometric product. Given two vectors\footnote{Points and circles  of Euclidean geometry are represented in upper case italic letters while its vectors in lower case bold letters. Among CGA elements, multivectors are represented in upper case bold italic letters while vectors are represented in lower case bold italic letters.}, $\mv{a}$ and $\mv{b}$, their geometric product is defined as
 \begin{align} \label{eq:gp}
     \mv{a}\mv{b} = \mv{a} \cdot \mv{b} + \mv{a} \wedge \mv{b}
 \end{align}
where $\mv{a} \cdot \mv{b}$ is the inner product that yields a scalar and $\mv{a} \wedge \mv{b}$ is the outer product that gives a bivector. They follow the properties  $\mv{a} \cdot \mv{b} = \mv{b} \cdot \mv{a}$ and $\mv{a} \wedge \mv{b} = - \mv{b} \wedge \mv{a}$.

Orthogonality of the basis vectors implies that $\e{i} \cdot \e{j} = 0, \ i \neq j$. 
\begin{align}\label{eq:geopro}
    \e{i}\e{j} = \e{i} \cdot \e{j} + \e{i} \wedge \e{j} = 
    \begin{cases}
        \pm 1, & i=j\\
        \e{i} \wedge \e{j}, \ \mbox{denoted as } \e{ij},  & i \neq j 
    \end{cases}
\end{align}
In this algebra, a basis change is done by introducing two null vectors $\e{0}$ and $\e{\infty}$ that represent a point at the origin and a point at infinity respectively:
\begin{align}
    \e{\infty} = \e{4} + \e{5} \ ; \  \e{0} = \frac{1}{2}(\e{5} - \e{4}) \label{eq:infeq0}
\end{align}
From~\eqref{eq:geopro}, it follows that $\e{\infty}^2 = \e{0}^2 = 0$ and $\e{\infty} \cdot \e{0} = -1$. 
$\mathbb{G}_{4,1}$ has $2^5 = 32$ basis elements called blades listed in Table~I in~\cite{Tobias2023}. %
A multivector $\mv{A}$ is a linear combination of those basis elements $\{ 1, \e{0}, \e{1}, \ldots, \e{0123\infty} \} $. 
The conformal pseudoscalar is $\mv{I} = \e{0123\infty}$. Therefore, the conformal dual of a multivector $\mv{A}$ is 
\begin{align}\label{eq:dual}
    \mv{A}^* = \mv{A}\mv{I}^{-1}, \ \mv{I}^{-1} = \e{0} \mv{I}_3^{-1} \e{\infty} 
\end{align}
where $\mv{I}_3 = \e{1}\e{2}\e{3}$ and $\mv{I}_3^{-1} = \e{3}\e{2}\e{1}$.

Here we use the formulation in \cite{Dorst2009} to represent geometric objects and their duals. Those objects are in general null space representations with respect to either the inner (IPNS) or the outer (OPNS) product~\cite{Tobias2023}.
An Euclidean point $\mathbf{x}\in \mathbb{R}^3$ is embedded as a null vector in CGA using the $\mbox{up}()$ function:
\begin{align}\label{eq:point}
    \mv{x} = \mbox{up}(\mathbf{x}) = \e{0} + \mathbf{x} + \frac{1}{2}\mathbf{x}^2\e{\infty}, \, \mv{x} \cdot \mv{x} = 0
\end{align}


OPNS is also known as the primal or direct representation of the geometric primitives and IPNS its dual according to~\eqref{eq:dual}. For instance, the direct representation of a sphere in CGA is given by $\mv{S} = \mv{p}_1 \wedge \mv{p}_2 \wedge \mv{p}_3 \wedge \mv{p}_4$
where $\mv{p}_1, \mv{p}_2, \mv{p}_3, \mv{p}_4$ are conformal representations of points on the sphere that are not all in the same plane. The dual representation is given by $\mv{S}^* = \mv{p}_S - \frac{1}{2} r^2 \e{\infty}$, where $\mv{p}_S$ is the conformal center point and $r$ is the radius of the sphere. 
These representations allow us to build CGA elements at our convenience, which is especially beneficial in kinematic analysis.
\begin{table}[!t]
\centering
 \begin{tabular}{|c | c | c|} 
\hline
CGA primitives & Direct/primal~(OPNS) & Dual~(OPNS) \\ 
\hline
Point $\mv{p}$ & $\mv{x}$ in~\eqref{eq:point}  & $\mv{x}$  \\ 
Point pair $\mv{A}$ &  $\mv{p}_1 \wedge \mv{p}_2$ & $\mv{S}_1 \vee \mv{S}_2 \vee \mv{S}_3$  \\ 
Sphere $\mv{S}$ &  $\mv{p}_1 \wedge \mv{p}_2 \wedge \mv{p}_3 \wedge \mv{p}_4$ & $\mv{p}_S - \frac{1}{2} r^2 \e{\infty}$  \\ 
Plane $\mv{E}$ &  $\mv{p}_1 \wedge \mv{p}_2 \wedge \mv{p}_3 \wedge \e{\infty}$ & $\mv{n} + d \e{\infty}$  \\ 
Line $\mv{L}$ &  $\mv{p}_1 \wedge \mv{p}_2 \wedge \e{\infty} $ & $\mv{E}_1 \vee \mv{E}_2$  \\ 
Circle $\mv{C}$ &  $\mv{p}_1 \wedge \mv{p}_2 \wedge \mv{p}_3$ & $\mv{S}_1 \vee \mv{S}_2$ or $\mv{S}_1 \vee \mv{E}_1$  \\ 
\hline
\end{tabular}
\caption{Geometric primitives in 3D CGA. The outer product `$\wedge$' acts as the join of different elements, its dual is the regressive product represented by `$\vee$'. $\mv{p}_S$ is the center of the sphere and $r$ is its radius; $\mv{n}$ is the normal vector of the plane and $d$ is the distance from the origin.}
\label{table:objects}
\end{table}%
Table~\ref{table:objects} lists the direct and dual representations of all CGA primitives. Note that to represent a plane, the basis elements of the normal vector $\mv{n}$ can only be $\e{1}$, $\e{2}$ and $\e{3}$ since it just represents a direction and hence we have to forego the $\e{0}$ and $\e{\infty}$ elements that constrain it. To obtain this vector from a given line, a grade 3 element, 
we have to keep only coefficients of $\e{145}$, $\e{245}$ and $\e{345}$ but replace the basis vectors by $\e{1}$, $\e{2}$ and $\e{3}$, respectively. We call this change of basis as `vectorizing' a line.\\
Translation along a vector $\mv{n}$, with a distance $r$ is defined by the following versor~\cite[Section 13.2.2]{Dorst2009}:
\begin{align} \label{eq:Tversor}    
    \mv{R}_\infty(r, \mv{n}) = 1 - \frac{r}{2}  (\mv{n} \wedge \e{\infty})
\end{align}
Rotations in CGA are given by the following versor~\cite[Sections 7.2 and 13.2.2]{Dorst2009}:
\begin{align} \label{eq:Rversor} 
    \mv{R}(\phi, \mv{B}) = \exp^{-\mv{B} \frac{\phi}{2}} = \cos \left( \frac{\phi}{2} \right) -  \sin \left( \frac{\phi}{2} \right) \mv{B}
\end{align}
Transformation between frames is accomplished using the following motors:
\begin{equation}
    \mv{M}(\theta) = \mv{R}(\theta, \e{12}), \,\,\mv{G}(a, d, \alpha) = \mv{R}_\infty(d, \e{3})\mv{R}(\alpha, \e{23})\mv{R}_\infty(a, \e{1}),
    \label{eq:mg_trans}
\end{equation}
where $a, d, \alpha$ are Denavit-Hartenberg (D-H) parameters as annotated in Fig.~\ref{fig:ex_3R}.
The angle between any two given lines, $\mv{L}_1, \mv{L}_2$, in CGA can be calculated as:
\begin{equation}
    \cos(\theta) = \frac{\mv{L}_{1} \cdot \mv{L}_{2}}{\sqrt{\mv{L}_{1} \cdot \mv{L}_{1}} \sqrt{\mv{L}_{2} \cdot \mv{L}_{2}}} 
    \label{eq:cos_angle}
\end{equation}

%% file: sections/3_kinematic_model.tex
\section{Inverse kinematic model}
\label{sec:ik_model}
\begin{figure}[!t]
    \centering
    \includegraphics[width=0.8\textwidth]{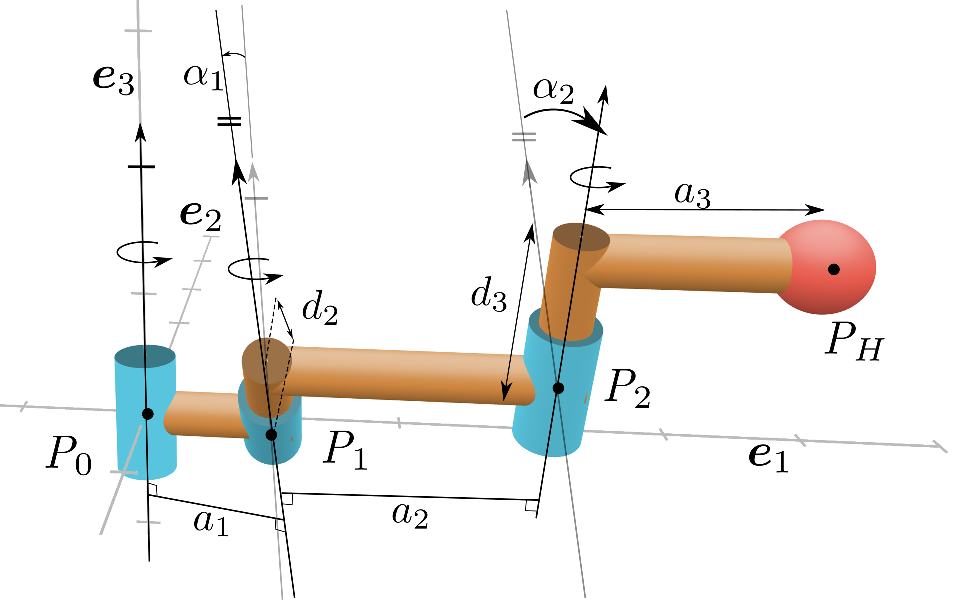}
    \caption{The schematic of a generic 3R robot and annotations relevant to CGA}
    \label{fig:ex_3R}
\end{figure}
A generic positional 3R robot is shown in Fig.~\ref{fig:ex_3R}. $P_i$, $i=\{0, 1, 2, H\}$ are points where the links are connected using revolute joints. $P_0$ is the origin and $P_H$ is where the end-effector lies when the robot is said to be in its `Home' position with joint angles $\theta_i = 0, \ i=\{1, 2, 3\}$. Inverse kinematics involves finding $\theta_i$ of the end-effector, given an arbitrary position, $P = (x, y, z)$.

We will follow Selig's inverse kinematics approach for 3R robots~\cite[Section 5.2]{selig}. In this approach, the necessary joint angles that move the end-effector from the home position $P_H$ to a target position $P$ are obtained by solving the equations:
\begin{equation}
    e^{\theta_1 S_1} e^{\theta_2 S_2} e^{\theta_3 S_3} \left( \begin{array}{cc}
        \mathbf{p}_H \\
         1 
    \end{array} \right) = \left( \begin{array}{cc}
        \mathbf{p} \\
         1 
    \end{array} \right)
    \end{equation}
where $\mathbf{p}_H$ and $\mathbf{p}$ are Euclidean vectors representing $P_H$ and $P$, $S_i, \ i=\{1, 2, 3\}$ is the Lie algebra element representing the $i$th joint. The equations are further written as:
\begin{equation}
   e^{\theta_2 S_2}  \left( \begin{array}{cc} \mathbf{a} \\ 1 \end{array} \right) = \left( \begin{array}{cc} \mathbf{b} \\ 1 \end{array} \right) \ 
   \mbox{with} \ 
   \left( \begin{array}{cc} \mathbf{a} \\ 1 \end{array} \right) = e^{\theta_3 S_3} \left( \begin{array}{cc} \mathbf{p}_H \\ 1 \end{array} \right) \ 
   \mbox{and} \ 
   \left( \begin{array}{cc} \mathbf{b} \\ 1 \end{array} \right) = e^{-\theta_1 S_1} \left( \begin{array}{cc} \mathbf{p} \\ 1 \end{array} \right) \label{eq:ik_ab}
\end{equation} 
where $\mathbf{a}$ and $\mathbf{b}$ are vectors representing points on circles parametrized by the first and last joint angles. Selig continues with an algebraic approach to solve these equations. The geometric interpretation is stated as finding the intersection of a pair of conics expressed in terms of $\theta_1$ or $\theta_3$, similar to Pieper's interpretation of IKM \cite{pieper_kinematics_1968}. 
However, CGA allows a better geometric intuition on how these solutions might be represented in the workspace by leveraging the fact that interactions between circles and their rotors are easy in CGA. We show that the inverse kinematic analysis reduces to finding the intersection between a fixed and a rotating circle.

In~\eqref{eq:ik_ab}, let the circles on which $\mathbf{a}$ and $\mathbf{b}$ lie be named $C_A$ and $C_B$. In $\mathbb{G}_{4,1}$, they are calculated as follows. The center of $C_B$ lies on $z$-axis and contains $P$. As listed in Table~\ref{table:objects}, $C_B$ in CGA is represented as the intersection of a sphere and a plane:
\begin{align}
    \mv{C}_B &= \mv{S}_B \vee \mv{E}_B, \ \mbox{with} \  \mv{S}_B^* = \mv{p}_{\mv{S}_B} - \frac{1}{2} r_B^2 \e{\infty}  \ \mbox{and} \ \mv{E}_B^* = \mv{n}_B + d_B \e{\infty} \label{eq:C_B}
\end{align}
where $\mv{S}_B^*$ is the dual representation (cf. Table~\ref{table:objects}) of the sphere with center $\mv{p}_{\mv{S}_B} = \mv{p}_0 = \mbox{up}(\mathbf{p}_0)$ and radius $r_b = || \mathbf{p} ||$. Plane $\mv{E}_B^*$ is also dually represented (cf. Table~\ref{table:objects}) with its normal $\mv{n}_B = \e{3}$ being the vector representing the $z$-axis and $d_B = z$.\\
Similarly, $C_A$ is obtained as the intersection of the following sphere and plane although expressing it as a CGA element is more nuanced:
\begin{align}
    \mv{C}_A &= \mv{S}_A \vee \mv{E}_A, \ \mbox{with} \  \mv{S}_A^* = \mv{p}_{\mv{S}_A} - \frac{1}{2} r_A^2 \e{\infty} \ \mbox{and} \ \mv{E}_A^* = \mv{n}_A + d_A \e{\infty} \label{eq:C_A}
\end{align}
where the center of the dual sphere $\mv{S}_A^*$ is $\mv{p}_{\mv{S}_A} = \mv{p}_2 = \mbox{up}(\mathbf{p}_2)$ and its radius $r_A = \sqrt{a_3^2 + d_3^2}$. The plane $\mv{E}_A^*$ is again dually represented whose normal vector $\mv{n}_A$ denotes the third joint axis obtained by transforming the unit vector, $\e{3}$, using the motors defined in.~\eqref{eq:mg_trans}:
\begin{equation}
    \mv{n}_A = \mv{T}_3 \e{3} \mv{T}_3^{-1} \ \mbox{with} \ \mv{T}_3 = \mv{M}(0)\mv{G}(\alpha_1, a_1, d_1)\mv{M}(0)\mv{G}(\alpha_2, a_2, d_2). \label{eq:p_j3}
\end{equation}
Additionally, the distance of the plane from the origin $d_A = \mv{n}_A \cdot \mv{p}_0$. 

Solving~\eqref{eq:ik_ab} implies rotating $\mv{C}_A$ about the second joint axis by $\theta_2$ to meet $\mv{C}_B$. Applying a rotor $\mv{R}_2$ to it, we obtain the rotated circle  as $\mv{C}_{A_{\theta_2}} = \mv{R}_2 \mv{C}_A \mv{R}_2^{-1}$. $\mv{R}_2$ as shown in~\eqref{eq:Rversor} needs a bivector to represent the plane of rotation perpendicular to the rotation axis. 
This is determined by transforming the bivector $\e{12}$ using the following motors (cf.~\eqref{eq:mg_trans}):
\begin{equation}
    \mv{B}_{j_2} = \mv{T}_2 \e{12} \mv{T}_2^{-1} \ \mbox{with} \ \mv{T}_2 = \mv{M}(0)\mv{G}(\alpha_1, a_1, d_1). \label{eq:p_j2}
\end{equation}


Thus, from~\eqref{eq:Rversor}, $\mv{R}_2 =  \exp^{-\mv{B}_{j_2} \frac{\theta_2}{2}}$ and the rotated circle $\mv{C}_{A_{\theta_2}}$ is a function of trigonometric functions of $\theta_2$. Therefore the intersection $\mv{x} = \mv{C}_{A_{\theta_2}} \vee \mv{C}_B$ is also a function of $\theta_2$. For this intersection to be a single real point, $\mv{x} \cdot \mv{x} = 0$ 
must be satisfied~\cite{Lasenby2004}, leading to an univariate quartic polynomial in $t_2 = \tan\frac{\theta_2}{2}$.\\

\textbf{Calculating $\theta_1$}: For a chosen $\theta_2$ value, the corresponding point $\mv{x}$ on circles $\mv{C}_{A_{\theta_2}}$ and $\mv{C}_B$ can be determined. If we consider $\mv{C}_B$, it follows from~\eqref{eq:ik_ab} that $\theta_1$ is the angle made by the directed arc on $\mv{C}_B$ that connects $\mv{p}$ to the point $\mv{x}$. To calculate this angle, let us consider two lines, $\mv{L}_{11} = \mv{p}_0 \wedge \mv{p} \wedge \e{\infty}$ and $\mv{L}_{12} = \mv{p}_0 \wedge \mv{x} \wedge \e{\infty}$. By projecting these lines onto plane $\mv{E}_B$, we can find the angle between them using~\eqref{eq:cos_angle}. Finding $\arccos()$ of the above value yields $\theta_1$, however the sign of the obtained value depends on the sign of the bivector obtained by vectorizing lines $\mv{L}_{11}$ and $\mv{L}_{12}$ as explained in Section~\ref{sec:cga_basics} and taking the outer product of the resulting vectors $\mv{B}_1 = - \mv{v}_{\mv{L}_{11}} \wedge \mv{v}_{\mv{L}_{12}}$. This bivector represents the positive rotation about the first joint axis. The negative sign comes from~\eqref{eq:ik_ab}. The sign of $\theta_1$ obtained from~\eqref{eq:cos_angle} must be adjusted to match the sign of $\mv{B}_1$.\\~\\
\textbf{Calculating $\theta_3$}: Similar to the calculation of $\theta_1$, we look for two points on $\mv{C}_A$ whose connecting directed arc makes an angle $\theta_3$ about its center. One such point is $\mv{p}_H$, to find the other one, we have to rotate the obtained $\mv{x}$, currently on $\mv{C}_{A_{\theta_2}}$ back to $\mv{C}_A$ by its corresponding $\theta_2$: $\mv{x}_{-\theta_2} = \mv{R}_2^{-1} \mv{x} \mv{R}_2$ with $\mv{R}_2 = \exp^{-\mv{B}_{j_2} \frac{\theta_2}{2}}$. It follows from~\eqref{eq:ik_ab} that $\theta_3$ is the angle made by the directed arc on $\mv{C}_A$ that connects $\mv{x}_{-\theta_2}$ to the point $\mv{p}_H$. To calculate this angle, let us again consider two lines, $\mv{L}_{31} = \mv{p}_2 \wedge \mv{p}_H \wedge \e{\infty}$ and $\mv{L}_{32} = \mv{p}_2 \wedge \mv{x}_{-\theta_2} \wedge \e{\infty}$. By projecting these lines onto the plane $\mv{E}_A$, we can find the angle between them using~\eqref{eq:cos_angle}. In this case, the sign of $\theta_3$ depends on the sign of the bivector $\mv{B}_3 =  \mv{v}_{\mv{L}_{31}} \wedge \mv{v}_{\mv{L}_{32}}$, where $\mv{v}_{\mv{L}_{31}}$ and $\mv{v}_{\mv{L}_{32}}$ are obtained by vectorizing $\mv{L}_{31}$ and $\mv{L}_{32}$. $\mv{B}_3$ represents the positive rotation about the third joint axis. The sign of $\theta_3$ obtained from~\eqref{eq:cos_angle} must be adjusted to match the sign of $\mv{B}_3$.\\~\\
\begin{figure}[!t]
    \centering
    \includegraphics[width=0.8\textwidth]{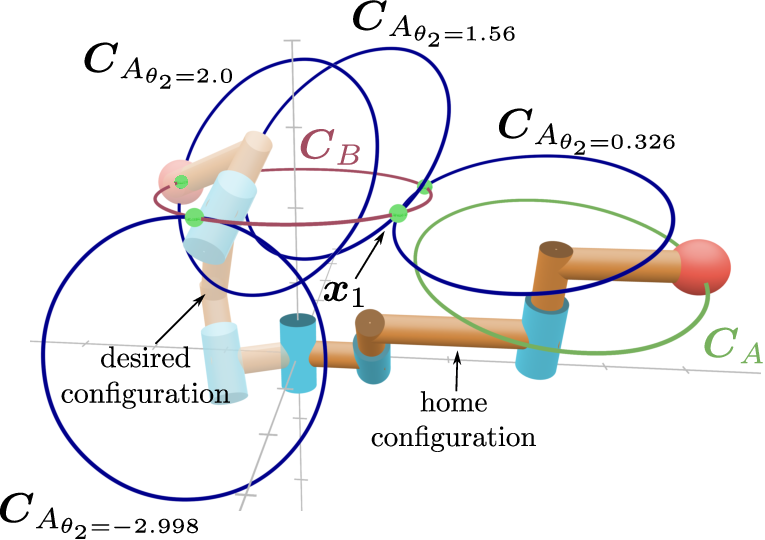}
    \caption{An example illustration of four IKS of a generic 3R robot interpreted as intersection between a fixed and rotating circle.}
    \label{fig:ik_circles}
\end{figure}
\textbf{Example}\\
Let us consider a generic 3R robot shown in Fig.~\ref{fig:ex_3R} with D-H parameters as $\mathbf{d} = [0, 1, 1], \mathbf{a} = [1, 2, 1.5], \mathbf{\alpha} = [\pi/4, -\pi/6, 0]$. We find its inverse kinematics solutions for a given arbitrary end-effector position: $\mathbf{p} = [-1.62, 0.465, 2.21]$. For demonstration purposes, $\mathbf{p}$ is chosen such that we have 4 IKS. In terms of CGA, $\mv{p} = \mbox{up}(\mathbf{p}) = -1.62 \e{1} + 0.465 \e{2} + 2.21 \e{3} + 3.36 \e{4} + 4.36 \e{5}$.
In home position, $ \mv{p}_0 = -0.5 \e{4} + 0.5 \e{5}, \ \mv{p}_1 = \e{1} + \e{5}, \ \mv{p}_2 = 3 \e{1} - 0.707 \e{2} + 0.707 \e{3} + 4.5 \e{4} + 5.5 \e{5}, \ \mv{p}_H = 4.5 \e{1} - 0.966 \e{2} + 1.67 \e{3} + 11.5 \e{4} + 12.5 \e{5}$
From~\eqref{eq:C_B}, 
\begin{flalign*}
    \mv{S}_B^* &= \mv{p}_0 - \frac{1}{2} 2.78^2 \e{\infty} \Rightarrow \mv{S}_B = -3.36 \e{1234} - 4.36 \e{1235} && \\
    \mv{E}_B^* &= \e{3} +  2.21 \e{\infty}  \Rightarrow \mv{E}_B = 2.21 \e{1234} + 2.21 \e{1235} - \e{1245} && \\
    \mv{C}_B &= \mv{S}_B \vee \mv{E}_B = 2.21 \e{123} + 3.36 \e{124} + 4.36 \e{125}
\end{flalign*}
From~\eqref{eq:C_A},
\begin{flalign*}
    &\mv{S}_A^* = \mv{p}_2 - \frac{1}{2} 1.802^2 \e{\infty} && \\ &\mv{S}_A = 3.87 \e{1234} + 2.87 \e{1235} - 0.707 \e{1245} - 0.707 \e{1345} - 3 \e{2345} && \\
& \mv{n}_A = -0.26 \e{2} + 0.96 \e{3} && \\
    &\mv{E}_A = \mv{n}_A +  (\mv{n}_A \cdot \mv{p}_H) \e{\infty} = 1.87 \e{1234} + 1.87 \e{1235} - 0.966 \e{1245} - 0.259 \e{1345} && \\
    &\mv{C}_A = \mv{S}_A \vee \mv{E}_A = 1.87 \e{123} - 2.42 \e{124} - 1.46 \e{125} + 0.317 \e{134} + 0.575 \e{135} && \\
    & \hspace{2.5cm} - 0.5 \e{145}  - 5.6 \e{234}  - 5.6 \e{235} - 2.9 \e{245} - 0.776 \e{345}
\end{flalign*}
from~\eqref{eq:p_j2}, $\mv{B}_{j_2} = 0.707 (\e{12} + \e{13} - \e{24} - \e{25} - \e{34} - \e{35})$.\\
The condition for the intersection of this circle with $\mv{C}_B$ to be a real point, $$
\mv{x} \cdot \mv{x} = -4.60\sin(\theta_2) - 0.95\sin(2\theta_2) + 1.09 \cos(\theta_2) - 1.99 \cos(2\theta_2) + 2.61 = 0$$
Solving the above equation gives four solutions, $\theta_2 = \{2.0, 0.326, 1.56, -2.998\}$.\\
Fig.~\ref{fig:ik_circles} shows the four rotated circles corresponding to the obtained $\theta_2$ solutions.
If we consider one of those circles, $\mv{C}_{A_{\theta_2 = 1.56}}$, its intersection with $\mv{C}_B$ gives:
\begin{flalign*}
\mv{x}_1 = \mv{C}_{A_{\theta_2 = 1.56}} \vee \mv{C}_B = 1.3 \e{1} - 1.07 \e{2} + 2.21 \e{3} + 3.36 \e{4} + 4.36 \e{5} &&
\end{flalign*}
Note that in $\mv{x}_1$ the coefficient of $\e{3}$ matches that of $\mv{p}$ as both of them lie in the plane of circle $\mv{C}_B$. In fact, this should hold for all intersection points of $\mv{C}_{A_{\theta_2}}$ and $\mv{C}_B$.\\ 
To find $\theta_1$, we calculate the projections of $\mv{L}_{11}$ and $\mv{L}_{12}$ onto $\mv{E}_B$:
\begin{flalign*}
\mv{L}_{11_{\mv{E}_B}} &= - 3.58 \e{134} - 3.58 \e{135} + 1.62 \e{145} + 1.03 \e{234} + 1.03 \e{235} - 0.465 \e{245} && \\
\mv{L}_{12_{\mv{E}_B}} &= 2.87 \e{134} + 2.87 \e{135} - 1.3 \e{145} - 2.37 \e{234} - 2.37 \e{235} + 1.07 \e{245}
\end{flalign*}
From~\eqref{eq:cos_angle}, $\theta_1$ could be $\pm 2.731$. Its sign is determined by calculating the bivector $\mv{B}_1 =  -1.13 \e{12}$, which is negative. Therefore $\theta_1 = -2.731$.
To find $\theta_3$, we calculate the projections of $\mv{L}_{31}$ and $\mv{L}_{32}$ onto $\mv{E}_A$:
\begin{flalign*}
\mv{L}_{31_{\mv{E}_A}} &= - 1.45 \e{124} - 1.45 \e{125} + 2.51 \e{134} + 2.51 \e{135} - 1.5 \e{145} && \\
\mv{L}_{32_{\mv{E}_A}} &= 3.79 \e{124} + 3.79 \e{125} - 1.29 \e{134} - 1.29 \e{135} + 1.19 \e{145} - 1.7 \e{234} && \\ 
& \hspace{0.4cm} - 1.7 \e{235} + 0.881 \e{245} + 0.236 \e{345}
\end{flalign*}
From~\eqref{eq:cos_angle}, $\theta_3$ could be $\pm 2.488$. Its sign is determined by calculating the bivector $\mv{B}_3 =  -1.32 \e{12} -0.354 \e{13}$, which is negative. Therefore $\theta_3 = -2.488$.
Following this procedure for the remaining $\theta_2$ solutions, the complete set of IKS for the given problem is $(0.0, 2.0, 1.0)$, $(2.58, 0.326, 2.138)$, $(-2.731, 1.56, -2.488)$, $(-1.341, -2.998, -1.753)$.

%% file: sections/4_conclusions.tex
\section{Conclusions}
A generic IKM for 3R robots was presented in this article as the intersection of two circles as elements of Conformal Geometric Algebra. CGA also allows us to extend the motion description to include orientations. This fact will be used to study the IKM for a generic 6R robot and its kinematic properties.
